\documentclass{article}

\PassOptionsToPackage{numbers, compress}{natbib}
\usepackage[preprint]{neurips_2025}

\usepackage[utf8]{inputenc} 
\usepackage[T1]{fontenc}    
\usepackage{hyperref}       
\usepackage{url}            
\usepackage{booktabs}       
\usepackage{amsfonts}       
\usepackage{nicefrac}       
\usepackage{microtype}      
\usepackage{xcolor}         
\usepackage{amsmath}
\usepackage[figurename=Figure]{caption}
\usepackage{natbib}
\usepackage{graphicx}  
\usepackage{subfigure}
\usepackage{cleveref}

\title{MASTER: Enhancing Large Language Model via Multi-Agent Simulated Teaching}

\author{Liang Yue$^*$, Yihong Tang$^*$, Kehai Chen$^\dag$, Jie Liu, Min Zhang
\\
Harbin Institute of Technology, Shenzhen, China \\
\texttt{\{220110921@stu.hit.edu.cn,  chenkehai@hit.edu.cn\}}
}

\begin{document}

\maketitle

\begingroup\def\thefootnote{*}\footnotetext{Equal contribution.}\endgroup
\begingroup\def\thefootnote{\dag}\footnotetext{Corresponding author.}\endgroup

\begin{abstract}
Instruction fine-tuning is crucial in NLP tasks, enhancing pretrained models' instruction-following capabilities and task-specific performance. However, obtaining high-quality fine-tuning data for large models is challenging due to data collection difficulties and high production costs. To address this, we propose MASTER, a novel data augmentation method that enriches original data through interactions among multiple agents with varying cognitive levels. We simulate three pedagogically grounded teaching scenarios, leveraging multi-agent conversations to generate high-quality teacher-student interaction data. 
Utilizing MASTER, we construct BOOST-QA, a fine-tuning dataset augmented from existing datasets like Orca-Math-200k, ProcQA, and OpenHermes2.5. 
Experiments show that models fine-tuned with BOOST-QA perform excellently across multiple benchmarks, demonstrating strong multitask generalization. Notably, MASTER significantly improves models' reasoning abilities in complex tasks, providing valuable insights for future research.
\end{abstract}

\section{Introduction}
\label{sec:introduction}

In recent years, instruction-tuning or post-training has become one of the cornerstones of large language models (LLMs)~\cite{han2024parameter, ouyang2022training}.
To meet the growing demand for data, data synthesis has been widely studied.
For example, Yoo et al. combined subsets of training examples and embedded them into model prompts to generate new, high-quality textual instances \cite{yoo2021gpt3mix}, demonstrating notable performance improvements in text classification tasks.
Ding et al. and Xu et al. proposed self-chat approaches based on predefined prompt templates and dialogue seeds, using large language models (LLMs) to generate diverse instruction data through self-dialogue \cite{ding2023enhancing, xu2023baize}.
For example, Yoo et al. combined subsets of training examples and embedded them into model prompts to generate new, high-quality textual instances \cite{yoo2021gpt3mix}, demonstrating notable performance improvements in text classification tasks.
While these methods enhance the diversity of synthetic instruction data, they rely heavily on manually crafted prompts and dialogue seeds and often lack clear interaction mechanisms, ultimately leading to a mismatch between the generated data and real-world instruction scenarios.

To address these challenges, our study introduces a novel multi-agent interaction framework aimed at enhancing original problem-solving datasets within simulated educational scenarios. Specifically, we developed a system comprising teacher and student agents. Through their collaborative interactions, any existing instruction dataset can be transformed into a \textbf{M}ulti-\textbf{A}gent \textbf{S}imulated \textbf{T}eaching \textbf{E}nhanced \textbf{R}esource (MASTER) framework. MASTER simulates three educational scenarios—error correction, collaborative debate, and analogical reasoning—by leveraging distinct conversational protocols and model prompts, ultimately resulting in the creation of a high-quality instruction dataset named BOOST-QA (Behaviorally Oriented Overlay of Simulated Teaching for QA).

Using the large-scale, high-quality instruction dataset BOOST-QA generated through the MASTER framework, we fine-tuned several mainstream base models.
To rigorously assess the effectiveness of the MASTER method, we conducted comprehensive experiments comparing the performance of base models fine-tuned on the original datasets, datasets augmented by other methods, and those created through MASTER. The results show that BOOST-QA significantly enhances the diverse capabilities of large language models (LLMs), outperforming several existing approaches focused on data augmentation and selection.

Our main contributions are summarized as follows:

\begin{itemize}
\item We introduce the application of multi-agent simulated instructional scenarios in post-training data synthesis and propose a novel data augmentation method, MASTER.

\item By applying MASTER to portions of Orca-math-200k, ProcQA, and OpenHermes2.5, we construct an efficient instruction fine-tuning dataset, BOOST-QA.

\item We design comprehensive experiments to assess our MASTER method. Notably, in benchmark tests, multiple models fine-tuned with 19K instruction-response pairs from our BOOST-QA exhibit significant performance improvements across various task domains.
\end{itemize}

\section{ Relate Work}
\label{sec:relate work}

\subsection{Data Synthesis and Augmentation}
In recent years, data synthesis and augmentation techniques have become essential for enhancing LLM performance and generalization capabilities \cite{li2022data, qu2020coda}. Early approaches focused primarily on simple lexical and positional transformations of raw data or employed LLMs to generate new training samples through synonym substitution of sentences from original datasets \cite{wei2019eda, shorten2021text, zhou2024pga}. While these methods partially addressed data scarcity, they risked introducing noise-induced semantic drift and often produced insufficient sample diversity for complex tasks \cite{zhou2021flipda}. To overcome these limitations, prompt-guided LLM approaches for comprehensive data expansion have emerged as a promising alternative, including methods such as constraint-augmented problem evolution to deepen original questions \cite{xu2023wizardlm}, multi-task contextual generation by sampling from seed pools \cite{wang2022self}, and knowledge-tree recursive QA to extend initial keywords \cite{cao2025condor}. Although these techniques improve content diversity and generalizability of synthetic instructions, they remain overly reliant on predefined prompts and keywords while lacking authentic natural language contexts. In contrast to prior work, our approach enhances problem-solving data quality by simulating real classroom learning scenarios through the incorporation of authentic educational events, achieving superior generalizability and ecological validity.

\subsection{Multi-Agent Simulation of Human Interaction}
Multi-agent simulation of human interaction has demonstrated significant potential in tasks such as personality analysis and social behavior research \cite{zhang2023exploring, wang2024investigating}. Early multi-agent systems primarily focused on two problem categories: goal-aligned collaborative tasks and game-theoretic competitive scenarios \cite{liu2022prospects, bo2024reflective, zhao2023competeai}. Recent advancements have substantially expanded agent populations to investigate social dynamics, exemplified by Mou et al.'s work employing agent swarms to model opinion propagation in social networks \cite{mou2024unveiling} and Stanford's "Virtual Town" simulating complex human behavioral patterns through agent socialization \cite{park2023generative}. However, practical applications leveraging multi-agent interaction for real-world problem-solving remain limited. Our proposed school-agent framework is specifically designed for synthesizing diverse, high-quality data to address this gap and meet operational requirements.

\subsection{Knowledge Distillation}
Knowledge distillation enables the transfer of knowledge from large teacher models to compact student models while preserving performance and reducing computational complexity. The seminal work by Hinton et al. first introduced this concept, demonstrating how soft labels could effectively compress models by distilling integrated knowledge from larger architectures \cite{hinton2015distilling}. Subsequent advancements by Jing et al. incorporated conditional generative adversarial networks to refine student logits outputs through adversarial training, achieving closer alignment with teacher outputs \cite{xu2017training}. Zhang et al. expanded the paradigm through mutual learning among multiple student models, proving effective for collaborative training scenarios \cite{zhang2018deep}. Novel directions emerged through Tung/Park et al.'s focus on relational similarity between teacher-student networks \cite{tung2019similarity, park2019relational}, while Xu et al.'s alternating sampling method significantly narrowed performance gaps in complex mathematical reasoning tasks \cite{xu2024speculative}. The recent "Branch-Merge Distillation" by DeepSeek successfully transferred DeepSeek-R1's capabilities to Qwen models across STEM benchmarks \cite{sun2025tinyr1}. Our approach diverges by employing agent-mediated interaction to inject novel cognitive patterns into raw data, fundamentally overcoming knowledge distillation's inherent limitations in generalization capacity and teacher-dependent behavioral constraints through enhanced data format learnability.

\section{Methodology}
\label{sec:methodology}

\subsection{Overview}
\label{sub:overview}
This section introduces our multi-agent classroom simulator, MACLASS. As illustrated in the figure, MACLASS enables LLMs to realistically play the roles of teachers and students through carefully designed prompts and uses a set of original question–answer data as input to generate simulated teaching interaction scenarios. MACLASS integrates diverse real-world educational settings and adheres to the design principle of embedding authentic and effective pedagogical methods into the multi-agent interaction process. Our approach primarily addresses the following two challenges: (1) How can effective educational principles be incorporated into agent-based teaching processes? (2) How can we ensure coherent and natural interactions among agents across different scenarios?

To address the first challenge, we integrate established pedagogical principles into classroom interactions through three key approaches: teachers guide students to correct mistakes and solve problems independently, facilitating experiential learning from errors \cite{heemsoth2016secondary}; teachers facilitate group debates to enhance critical thinking and analytical skills \cite{darby2007debate}; and teachers encourage analogical reasoning, enabling students to develop inductive learning abilities by solving structurally similar problems \cite{chen1999schema}. The above methods are each developed into distinct instructional scenarios, featuring multi-turn interactions among multiple agents. The utterances from these agents are then concatenated and organized into ShareGPT-format data, effectively integrating foundational educational principles with the reasoning capabilities of intelligent agents within the dataset.

To address the second issue, we ensure that the interactions between multiple agent roles are coherently and logically controlled. To this end, we design precise interaction management rules that govern the speaking order of agents across different teaching scenarios, and assign scenario-specific prompts to each agent at different turns. Specifically, we model the dialogue process as: \(\mathcal{D} = [\mathcal{D}_1, \mathcal{D}_2, \ldots, \mathcal{D}_n, \rho]\)
 ,where $D_i$ denote the utterance content of an agent at the $i$-th turn, and let $\rho$ represent the prompt used by the agent at the current teaching step. Under this framework, the predefined speaking order and prompt assignment strategy ensure the coherence of the dialogue flow, effectively achieving the intended pedagogical goals.

By adopting this method, we implement a fluent and well-controlled multi-agent classroom interaction module that successfully simulates authentic and effective teaching processes. The detailed configuration is provided in the Appendix.

\subsection{Agent role construction}
\label{sub:agent}
Studies have shown that interactive teaching can significantly enhance students' learning quality. However, its effective implementation relies on efficient dialogue exchanges between teachers and students, which remains a challenging task \cite{dorimana2022teacher}. Simulating high-quality classroom interactions using multiple agents often encounters problems such as agents competing for roles, drifting off-topic, or redundantly repeating previous responses. To address these issues, we introduce two strictly constrained types of agents: Teacher agents and Student agents. In our framework, we assign large language models (LLMs) with distinct prompts that enable them to assume different roles in a multi-agent setting or to perform different instructional tasks within the same role definition. This approach facilitates both functional cooperation and procedural control. Formally, this can be expressed as \(\mathcal{A} = (\mathcal{L}, P_{R_{i}}), \quad 
R_{i} \in R = \left[ R_{1}, R_{2}, \ldots, R_{n} \right]\).

In our framework, each agent \(\mathcal{A}\) powered by an LLM \(\mathcal{L}\) is assigned role-specific prompts \(P_{R_{i}}\) corresponding to distinct task phases \(R_{i}\). For instance, a student agent utilizes different prompts when initially making an error versus when correcting it. This structured prompt design ensures agents operate efficiently within their designated roles and phases, minimizing role confusion and task redundancy, thereby enhancing the quality and stability of the collaboratively generated educational dialogues. 

\paragraph{Teacher Agents}
In a classroom environment, the teacher not only serves as the primary source of knowledge transmission and student guidance, but also plays a pivotal role in shaping the overall learning experience, managing instructional dynamics, and fostering critical thinking. As a core component of the multi-agent classroom framework, the teacher must simultaneously fulfill multiple instructional functions, including delivering content, assessing student understanding, providing timely feedback, and adapting pedagogical strategies to accommodate diverse learning needs.

Upon receiving an original question and its standard answer, the teacher agent conveys the problem details to the student, offering brief explanations to facilitate understanding. When provided with a student's solution, along with the corresponding question and standard answer, the teacher agent identifies any errors in the student's response and supplements the instruction with correct problem-solving strategies, guiding the student to independently rectify previous mistakes. This structured approach enhances the quality and stability of the educational dialogues generated through multi-agent collaboration.

\paragraph{Student Agents}

As the recipients of instructional content and the primary agents in the problem-solving process, students play a pivotal role in integrating the three pedagogical methods introduced in ~\ref{sub:overview}~ into realistic educational scenarios.Student agents are expected not only to respond to teacher instructions but also to revise their previous answers based on prior attempts and peer debates, articulating their own perspectives accordingly.

\subsection{Class specific settings}
\label{sub:class}
\begin{figure}[t]
  \centering
  \includegraphics[width=1.0\linewidth]{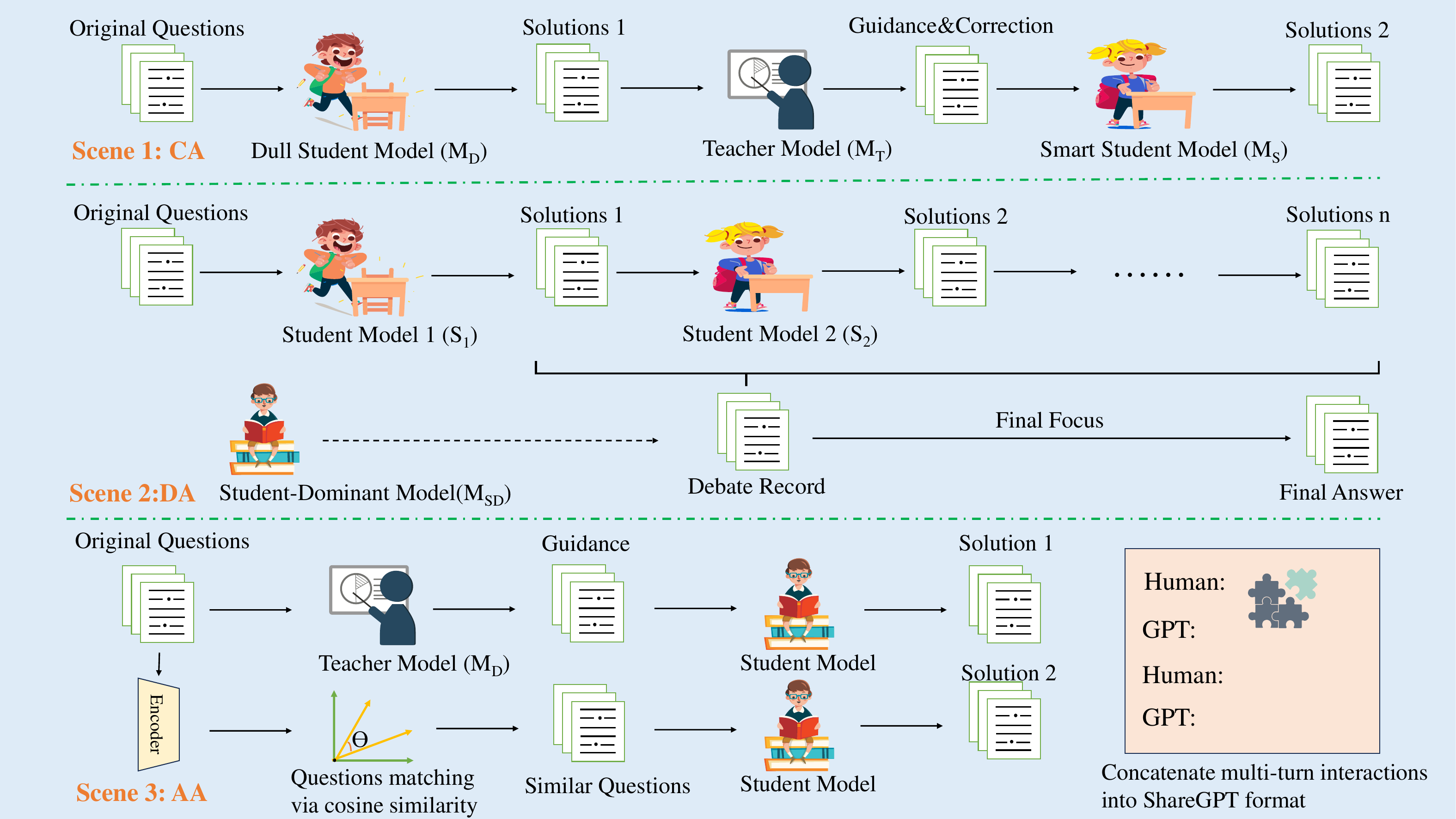}
  \caption{A multi-agent system-based data augmentation pipeline that simulates three different pedagogical contexts to enhance question-answering datasets. From top to bottom, the scenes are Correction Augmentation, Debatement Augmentation, and Analogical Augmentation.}
  \label{fig:data}
\end{figure}

Our work creates a classroom environment that diverges from traditional instructional paradigms. Multi-agent systems that rely solely on predefined operations are insufficient to effectively simulate concrete pedagogical strategies. To authentically integrate the pedagogical methods outlined in ~\ref{sub:overview}~ into the synthesized multi-agent classroom dialogue data, we have designed and developed a Classroom Interaction Manager comprising three modules: "Error Making and Correction," "Debate," and "Analogical Problem Retrieval and Solving." These modules respectively support the control of three distinct scenarios, as illustrated in \Cref{fig:data}. The multi-turn question-answering augmented data generated from these three scenarios are concatenated in the ShareGPT format to construct a high-quality dataset named BOOST-QA.

\paragraph{Error Correction Module}
This module is designed to simulate a classroom scenario where a student agent initially provides an incorrect or incomplete solution to a given problem. Subsequently, a teacher agent analyzes the student's response in conjunction with the standard answer to identify errors and offer correct reasoning. Finally, the student agent, leveraging the prior interaction and the standard answer, independently formulates a corrected solution. Specifically, we employ the Qwen2.5-0.5B-Instruct model with a temperature setting of 0.8 for the initial student response, promoting the generation of diverse and imperfect answers. For the subsequent correction phase, both teacher and student agents utilize the more capable Qwen2.5-14B-Instruct model with a temperature of 0.2, ensuring accurate error identification and high-quality reasoning.

In practice, training models with augmented data constructed by this module can inject structured noise into the gradient descent process. This simulates interactions across different cognitive levels, enabling the model to explore high-loss regions associated with student model errors and converge along smoother paths defined by corrected answers. Such an approach facilitates escaping local minima and aids in identifying and avoiding common error patterns during inference. The gradient expression is as follows:

\begin{equation}
\nabla_\theta \mathcal{L}_{\text{aug}}(\theta) \approx \nabla_\theta \mathcal{L}(\theta) + \gamma \cdot \mathbb{E}_{(x, \delta y) \sim \Delta \mathcal{D}} \left[ \ell\left(f_\theta(x), \delta y\right) \right].
\end{equation}

In this context, $\mathcal{L}(\theta)$ denotes the conventional supervised learning loss function. The parameter $\gamma$ represents the perturbation intensity introduced by multi-agent interactions, influenced by factors such as inference temperature. The term $\mathbb{E}_{(x, \delta y) \sim \Delta \mathcal{D}}$ signifies the expectation over the augmented data distribution $\Delta \mathcal{D}$. Here, $\ell$ is the per-sample loss function, $f_\theta(x)$ corresponds to the corrected solution generated by the 14B teacher model for the original problem $x$, and $\delta y$ denotes the perturbed answer label produced by the 0.5B student model.

\paragraph{Debate Interaction Module}

This module is designed to construct multiple student agents, each analyzing the question and the responses of other students to express their own perspectives. This approach aims to capture diverse problem-solving strategies and enhance data diversity.We employ identical prompts to build three student agents: $S_1$, $S_2$, and $S_3$. $S_3$ utilizes the Qwen2.5-14B-Instruct model with a temperature setting of 0.2, serving as the summarizer and decision-maker in the debate. Conversely, $S_1$ and $S_2$ employ the Qwen2.5-7B-Instruct model with a temperature of 0.6, acting as regular participants in the discussion. During the data augmentation process, $S_1$ and $S_2$ take turns speaking for one or two rounds, after which $S_3$ provides a summary. The output of the agent in the i-th round, denoted as $A_i$, can be represented by \(A_i = f(H, \theta_i)\), where $H$ signifies the current classroom dialogue history, and $theta_i$ denotes the output temperature corresponding to the agent.

Analogous to the error correction module, this component employs discussions among multiple agents to simulate the cognitive diversity of students. By fine-tuning the model with debate-enhanced data, it facilitates the smoothing of the loss landscape, thereby guiding the optimization trajectory away from sharp local extrema and enabling the learning of diverse, high-quality problem-solving strategies. The gradient representation of this module is as follows:

\begin{equation}
\nabla_{\theta}\mathcal{L}_{\text{aug}} = \mathbb{E}_{(x,y)\sim\mathcal{D}}\left[
\underbrace{\nabla_{\theta}\ell(f_{\theta}(x),y)}_{\text{original gradient}} + 
\lambda\underbrace{\sum_{k = 1}^{K}\nabla_{\theta}\ell(f_{\theta}(x),y_{k}^{*})}_{\text{debate gradient}}
\right].
\label{eq:main}
\end{equation}

Here, \(\mathbb{E}_{(x,y)\sim\mathcal{D}}\) denotes the expectation over the augmented data distribution \(\mathcal{D}\). The term \(\nabla_{\theta}\ell(f_{\theta}(x),y)\) represents the original gradient. The parameter \(\lambda\) is the debate intensity coefficient, influenced by factors such as model inference temperature and the number of debate rounds. \(K\)denotes the number of debate rounds, \(y_{k}^{*}\) is the answer provided by the agent in the current round, and \(\ell(f_{\theta}(x),y_{k}^{*})\) represents the single-sample loss.

\paragraph{Similar Question Retrieval Module}
This module facilitates basic interactions between teacher and student agents. After the student completes the first-round response, the system retrieves similar questions based on the initial prompt to construct analogy-based data by having the student solve a new, related problem. This approach encourages the application of existing knowledge to analogous scenarios, enhancing adaptability.

Specifically, a small subset of questions is randomly selected from the original dataset as the first-round questions \(Q_{1st}\), while the remaining questions form the retrieval pool. Using the all-MiniLM-L6-v2 model, embeddings are computed, and cosine similarity is employed to identify questions in the pool that closely resemble \(Q_{1st}\). From these, one question is randomly chosen as the analogy reasoning question \(Q_{2nd}\) for the student's second-round response, as illustrated in the accompanying formula. Both student and teacher agents utilize the Qwen2.5-14B-Instruct model with a temperature setting of 0.2. $Q$ denotes the complete question dataset, and $Q_{1st}^c$ represents the complement set of first-round questions, the analogy retrieval process is as follows:
\begin{equation}
Q_{2st} = \text{Random}\left(\text{Top-k}\left(\cos\_\text{sim}(Q_{1st}, Q_{1st}^c)\right)\right), \quad Q_{1st} \cup Q_{1st}^c = Q.
\label{eq:question_retrieval}
\end{equation}

Specifically, we employ the ShareGPT format to concatenate two rounds of question-answer dialogues from multi-agent students addressing similar problems into a single mixed training sample. This approach facilitates joint modeling of locally similar samples within the semantic space during model training, effectively serving as an implicit interpolation-based augmentation. Moreover, this mechanism encourages the model to focus on subtle differences between similar samples, thereby learning the core semantics of the task and enhancing robustness. The loss function for training the model with this augmented data is defined as follows:

\begin{equation}
\mathcal{L}_{\text{mix}} = -\log p(y_1, y_2 \mid x_1, x_2).
\end{equation}

Here, \(x_1\) and \(x_2\) denote the original problem texts, while \(y_1\) and \(y_2\) represent the corresponding original solution texts. The formula signifies the learning of the joint distribution between similar samples.
\section{Experiments}
\label{sec:experiments}
\subsection{Experimental Setup}
\label{sub:setup}
\paragraph{Training datasets.}

We utilized the following instruction-tuning datasets: (1) Orca-Math-Word-200K: A high-quality dataset of elementary school math question-answer pairs generated through multi-agent collaboration \cite{mitra2024orca}. (2) ProcQA: A dataset comprising mixed-modality programming question-answer pairs extracted from the StackOverflow community \cite{li2024procqa}. (3) OpenHermes 2.5: A general-purpose dataset encompassing tasks such as commonsense question answering and reasoning. We randomly selected 10,000 samples each from Orca-Math-Word-200K and ProcQA, and 9,000 samples from OpenHermes 2.5. These were combined to form the original training dataset, referred to as ori-data. Applying the MASTER data augmentation method to ori-data yielded an enhanced dataset of equal size (19,000 samples), termed BOOST-QA.
\paragraph{Evaluation datasets.}

We evaluated our method on HumanEval \cite{chen2021evaluating}, MBPP \cite{austin2021program}, MATH \cite{hendrycks2021measuring}, MMLU-PRO-MATH \cite{wang2024mmlu}, MMLU \cite{hendrycks2020measuring}, ARC \cite{clark2018think} and SCI-Q\cite{welbl2017crowdsourcing}.  These datasets encompass various domains and task types, including human-written coding challenges, mathematical problem-solving, multi-choice questions, and scientific reasoning, thereby providing a comprehensive assessment of our method's capabilities. During evaluation, we assessed the zero-shot capabilities of the MA-Gen model series across these datasets. The inference temperature was set to 0 for HumanEval and MBPP, and to 0.2 for all other datasets.

\paragraph{Models for data augmentation and training.}

We employed the Qwen2.5-Instruct series of models \cite{yang2024qwen2} as the foundational models to enhance the original data through multi-agent classroom interactions; specific configurations are detailed in ~\ref{sub:class}~. To evaluate the effectiveness of MASTER, we utilized three base models: LLaMA-3-7B-base \cite{grattafiori2024llama}, Qwen2.5-7B-base \cite{yang2024qwen2}, and Mistral-7B-base \cite{2023arXiv231006825J}. By fine-tuning these models on BOOST-QA, ori-data, and other high-quality datasets constructed using alternative methods, and subsequently comparing their performances, we validated the superiority of the MASTER method in enhancing data quality.

\subsection{Baselines}
\label{sub:baselines}
We selected four baseline methods for comparison with MASTER. First, we employed traditional text augmentation techniques by injecting character-level noise into the original text. Inspired by EDA, RandomAug and SpellingAug \cite{ma2019nlpaug} , were selected and are both open-sourced in the GitHub project nlpaug. 

The third baseline is TAGCOS \cite{zhang2024tagcos}, which computes gradient representations for each sample in the original dataset, clusters similar data points, and then applies a greedy algorithm within each cluster to effectively select high-quality data subsets for instruction fine-tuning. This approach emphasizes efficiency and relevance in data selection, aiming to reduce redundancy while preserving diversity in the fine-tuning corpus.

The final baseline is CoT Collection, proposed by Seungone Kim et al., which aims to enhance the reasoning capabilities of small and medium-sized language models in zero-shot and few-shot tasks through chain-of-thought (CoT) fine-tuning \cite{kim2023cot}. It aggregates a large and diverse set of CoT-annotated samples from multiple sources to provide explicit reasoning supervision, thereby helping models better learn intermediate reasoning steps.

To ensure a fair comparison, we used the TAGCOS method to select high-quality subsets of 5K, 5K, and 9K samples from the original training datasets Orca-Math-200K, ProcQA, and OpenHermes2.5, respectively, and combined them into a new high-quality training set. RandomAug, SpellingAug, CoT Collection, and our MASTER were then applied to randomly augment an equal amount of original data. Each method was used to fine-tune the pretrained model using LoRA for 2 epochs with a learning rate of 1e-4.

\subsection{Main Results}
\label{sub:main results}
We present the primary results of different models trained on BOOST-QA and Ori-Data across various benchmarks in \Cref{tab:ma-gen-comparison-part1} and \Cref{tab:ma-gen-comparison-part2}, and compare them with multiple baseline methods in \Cref{tab:baselines-comparison}. We use accuracy as the evaluation metric for mathematics and general tasks. For objective questions, answers are extracted using regular expressions, while subjective questions are evaluated for correctness by the Qwen2.5-14B-Instruct model based on the reference answers. For programming tasks such as HumanEval and MBPP, we adopt Pass@1 as the evaluation metric. Our findings are as follows:

\paragraph{BOOST-QA has demonstrated performance improvements across various models.}
In the experimental results presented in \Cref{tab:ma-gen-comparison-part1} and \Cref{tab:ma-gen-comparison-part2}, all baseline models fine-tuned with the augmented dataset BOOST-QA 
generally outperformed those fine-tuned with the original, unaugmented dataset Ori-Data across multiple benchmark tests. This indicates that our data augmentation method effectively enhances the learnability of the original data, thereby improving the model's generalization ability for the tasks.

\paragraph{The BOOST-QA dataset enabled models to achieve better performance compared to other baselines.}
In the experimental results presented in \Cref{tab:baselines-comparison}, the LLaMA3-8B-base model fine-tuned with BOOST-QA outperforms various baseline methods across multiple benchmarks in mathematics, programming, and general tasks. This highlights the advantages of our MASTER data augmentation method.

\begin{table*}[ht]
  \centering
  \caption{Performance comparison of models fine-tuned with Ori-Data and BOOST-QA (Part 1).}
  \label{tab:ma-gen-comparison-part1}
  \small
  \resizebox{\linewidth}{!}{
    \begin{tabular}{lcc cc cc cc}
      \toprule
      \textbf{Model} 
      & \multicolumn{2}{c}{\textbf{MATH}} 
      & \multicolumn{2}{c}{\textbf{MMLU-PRO-MATH}} 
      & \multicolumn{2}{c}{\textbf{MBPP}} 
      & \multicolumn{2}{c}{\textbf{HumanEval}} \\
      \cmidrule(lr){2-3} \cmidrule(lr){4-5} \cmidrule(lr){6-7} \cmidrule(lr){8-9}
      & Ori & BOOST-QA & Ori & BOOST-QA & Ori & BOOST-QA & Ori & BOOST-QA \\
      \midrule
      LLaMA3-8B      & 21.58 & \textbf{23.90} & 13.55 & \textbf{27.39} & 65.30 & \textbf{67.20} & 39.02 & \textbf{50.61} \\
      Qwen2.5-7B     & \textbf{71.00} & 70.54 & 24.35 & \textbf{44.41} & 78.00 & \textbf{79.10} & 22.56 & \textbf{42.07} \\
      Mistral-7B     & 15.74 & \textbf{17.58} & 7.18 & \textbf{13.92} & \textbf{56.30} & 55.00 & 17.68 & \textbf{28.05} \\
      \bottomrule
    \end{tabular}
  }
\end{table*}

\begin{table*}[ht]
  \centering
  \caption{Performance comparison of models fine-tuned with Ori-Data and BOOST-QA (Part 2).}
  \label{tab:ma-gen-comparison-part2}
  \small
  \resizebox{\linewidth}{!}{
    \begin{tabular}{lcc cc cc cc}
      \toprule
      \textbf{Model} 
      & \multicolumn{2}{c}{\textbf{MMLU}} 
      & \multicolumn{2}{c}{\textbf{ARC}} 
      & \multicolumn{2}{c}{\textbf{SCI-Q}} 
      & \multicolumn{2}{c}{\textbf{AVERAGE}} \\
      \cmidrule(lr){2-3} \cmidrule(lr){4-5} \cmidrule(lr){6-7} \cmidrule(lr){8-9}
      & Ori & BOOST-QA & Ori & BOOST-QA & Ori & BOOST-QA & Ori & BOOST-QA \\
      \midrule
      LLaMA3-8B      & 48.13 & \textbf{48.13} & 57.76 & \textbf{61.52} & 76.50 & \textbf{80.10} & 45.98 & \textbf{51.26} \\
      Qwen2.5-7B     & 24.05 & \textbf{50.12} & 20.90 & \textbf{68.52} & 20.00 & \textbf{69.70} & 37.27 & \textbf{60.64} \\
      Mistral-7B     & 27.59 & \textbf{35.89} & 32.25 & \textbf{47.70} & \textbf{70.50} & 49.10 & 32.46 & \textbf{35.32} \\
      \bottomrule
    \end{tabular}
  }
\end{table*}

\begin{table*}[ht]
  \centering
  \caption{Performance comparison of models fine-tuned with BOOST-QA and other baselines.}
  \label{tab:baselines-comparison}
  \small
  \resizebox{\linewidth}{!}{
    \begin{tabular}{lcc cc cc cc cc cc cc}
      \toprule
      \textbf{Method} & \textbf{Ori} & \textbf{RandomAug} & \textbf{SpellingAug} & \textbf{TAGCOS} & \textbf{CoT-fine} & \textbf{BOOST-QA} \\
      \midrule
      MATH        & 21.58 & 21.16 & 20.86 & \textbf{26.50} & 21.02 & 23.90 \\
      MMLU-PRO-MATH    & 13.55 & 14.58 & 13.40 & 12.95 & 14.36 &\textbf{27.39} \\
      MBPP   & 65.30 & 61.90 & 63.50 & 61.40 & 61.40 & \textbf{67.20} \\
      HumanEval   & 39.02 & 31.10 & 37.20 & 42.68 & 33.54 & \textbf{50.61} \\
      MMLU   & 48.13 & 38.35 & 24.21 & 46.91 & 41.08 & \textbf{48.13} \\
      ARC   & 57.76 & 41.98 & 22.61 & 61.09 & 47.87 & \textbf{61.52} \\
      SCI-Q   & 76.50 & 62.30 & 22.70 & \textbf{84.00} & 68.50 & 80.10 \\
      Average   & 45.98 & 38.78 & 29.21 & 47.93 & 41.11 & \textbf{51.26} \\
      \bottomrule
    \end{tabular}
  }
\end{table*}

\begin{figure}[htbp]
  \centering
  \includegraphics[width=0.9\linewidth]{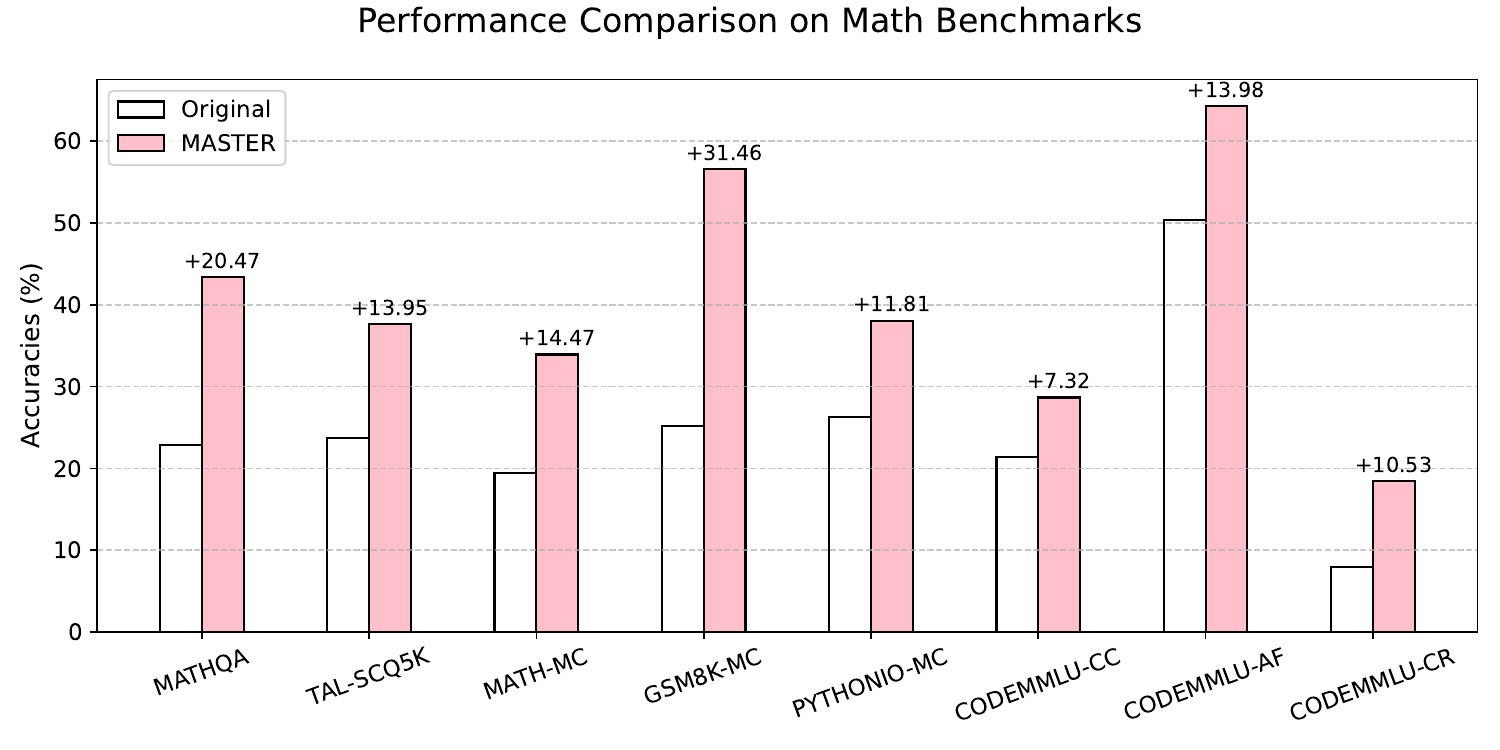}
  \caption{Results from the complex multiple-choice question test show a maximum improvement of 31.46\% and an average improvement of 15.50\%. CODEMMLU-CC, CODEMMLU-AF, and CODEMMLU-CR are abbreviations for CODEMMLU-CODE-COMPLETION, CODEMMLU-API-FRAMEWORKS, and CODEMMLU-CODE-REPAIR, respectively.}
  \label{fig:data2}
\end{figure}

\paragraph{The BOOST-QA dataset significantly enhances the model's capability in solving complex multiple-choice questions.}
 Notably, We found that our MASTER series models achieved remarkable improvements on multiple-choice question tasks. Upon observing this phenomenon, we incorporated additional multiple-choice benchmark tests, with the results illustrated in \Cref{fig:data2}. Across the current eight test datasets, the MASTER method demonstrated consistent improvements exceeding 5\% compared to training with the original data, peaking at a 31.46\% enhancement. Through analysis of the models' inference outputs, we found that the MASTER-series models generated significantly longer reasoning chains than those trained on the original data, where outputs tended to be shorter or limited to direct option selection. This indicates that our BOOST-QA dataset effectively internalizes the models' reasoning capabilities, enabling better generalization when addressing complex problems.

\graphicspath{{figs/}}  

\subsection{Ablation Study}
\label{sub:ablation study}
In this section, we design ablation experiments by fine-tuning the LLaMA3-8B-base model using training data constructed from various combinations of educational scenarios. This approach aims to assess the impact of different teaching scenarios on model performance. Additionally, we explore the distributions of augmented and original data in the hidden layers after applying t-SNE dimensionality reduction.
\begin{table*}[ht]
  \centering
  \caption{ Performance comparison of models fine-tuned with partial and full multi-agent simulated teaching scenarios across multiple benchmarks. ME, DB, and EP are the abbreviations for Make Error, Debate, and Expand, respectively.}
  \label{tab:Ablation of Teaching Scenarios}
  \small
  \resizebox{\linewidth}{!}{
    \begin{tabular}{lcc cc cc cc cc }
      \toprule
      \textbf{Method} & \textbf{Ori} & \textbf{\text{ME\&DB}} & \textbf{\text{ME\&EP}} & \textbf{\text{DB\&EP}} & \textbf{ME} & \textbf{DB} & \textbf{EP} & \textbf{Full} \\
      \midrule
      MATH        & 21.58 & 19.88 & 19.90 & 23.48 & 17.96 & 21.90 & 22.40 & \textbf{23.90} \\
      MMLU-PRO-MATH    & 13.55 & 16.14 & 17.25 & 17.62 & 13.40 & 11.32 & 10.58 & \textbf{27.39} \\
      MBPP   & 65.30 & 56.10 & 52.90 & 66.10 & 52.90 & 63.50 & 61.60 & \textbf{67.20} \\
      HumanEval   & 39.02 & 31.70 & 36.59 & 45.73 & 23.78 & 47.56 & 44.51 & \textbf{50.61} \\
      MMLU   & 48.13 & 40.08 & 35.36 & 32.64 & 40.54 & 19.86 & 31.40 & \textbf{48.13} \\
      Average   & 37.52 & 32.78 & 32.40 & 37.11 & 29.72 & 33.19 & 34.10 & \textbf{43.45} \\
      \bottomrule
    \end{tabular}
  }
\end{table*}
\paragraph{Effects of two scenarios}
In a subsequent series of ablation experiments, as detailed in \Cref{tab:Ablation of Teaching Scenarios}, we explored the impact of combining different pairs of educational scenarios on model performance. Specifically, ME\&DB (Make Error and Debate), ME\&EP (Make Error and Expand), and DB\&EP (Debate and Expand) denote models fine-tuned with augmented data that integrates these scenario pairs. To construct these dual-scenario datasets, we meticulously removed data lacking the corresponding scenarios from BOOST-QA and supplemented it with carefully selected samples from Ori-Data. The experimental results indicate that augmenting with only one or two scenarios fails to significantly enhance model performance. In contrast, the MASTER model, which is trained with data augmented from all three educational scenarios, consistently outperforms the model trained on the original data across all test sets. This finding underscores the complementary and indispensable roles of each educational scenario in the data augmentation process, highlighting that a combination of multiple scenarios is essential for optimal model performance.

\paragraph{Effects of one scenarios}
 In \Cref{tab:Ablation of Teaching Scenarios}, we systematically evaluated the performance of models trained on datasets augmented using only a single educational scenario—specifically, ME (Make Error), DB (Debate), or EP (Expand). The results consistently showed that these models underperformed compared to the model trained solely on the original dataset across all test sets. This outcome highlights the limitations of augmenting data with a single scenario, as it fails to provide the comprehensive learning experiences necessary for robust model performance. Furthermore, the models trained with only one scenario exhibited a lack of robustness when faced with diverse test conditions, indicating that a singular approach is insufficient for comprehensive learning. Our findings suggest that integrating multiple educational scenarios is crucial for enhancing the adaptability and generalizability of the models, as each scenario contributes unique learning signals that collectively improve model performance. Therefore, we conclude that a combination of varied educational scenarios is essential for effectively improving model performance, as it provides a more holistic and diverse learning environment.

\section{Conclusion}

This study systematically investigates the impact of constructing multi-agent instructional scenarios on question-answering (QA) data augmentation. To obtain high-quality instruction fine-tuning data, we simulate three distinct educational scenarios using multiple agents, introducing varying levels of cognitive interaction into the original data. This approach aims to enhance the convergence efficiency of the base model on the augmented data. Furthermore, during the inference phase, we implement an error-correction interaction pattern that mirrors the structure of the training data, ensuring consistency between training and reasoning processes. Our experimental results validate the effectiveness of this comprehensive framework in improving model performance.


\medskip

{
\small

\bibliographystyle{unsrtnat}
\bibliography{references}
}

\end{document}